\documentclass{article}
\usepackage{spconf,amsmath,graphicx}
\usepackage{amsfonts}
\usepackage{booktabs}
\usepackage{threeparttable}
\usepackage{enumerate}
\usepackage{subfigure,epsfig,amsfonts,amsmath,amssymb,graphics, latexsym, times,algorithm, algorithmic,bbm,cite,bm}
\usepackage{balance}
\usepackage{color}
\usepackage{comment}
\usepackage{enumitem}

\title{HDNet: Hierarchical Dynamic Network for Gait Recognition\\ using Millimeter-Wave Radar}
\name{\begin{tabular}{c}Yanyan Huang$^1$, Yong Wang$^{1,2}$, Kun Shi$^1$, Chaojie Gu$^1$, Yu Fu$^1$, Cheng Zhuo$^1$, Zhiguo Shi$^{1,2}$\sthanks{Corresponding author.}\end{tabular}}
\address{\normalsize$^{1}$ Zhejiang University, Hangzhou, China\\
\normalsize$^{2}$ Key Laboratory of Collaborative Sensing and Autonomous Unmanned Systems of Zhejiang Province, Hangzhou, China}

\begin{document}
%
\maketitle
%
\begin{abstract}
Gait recognition is widely used in diversified practical applications. Currently, the most prevalent approach is to recognize human gait from RGB images, owing to the progress of computer vision technologies. Nevertheless, the perception capability of RGB cameras deteriorates in rough circumstances, and visual surveillance may cause privacy invasion. Due to the robustness and non-invasive feature of millimeter wave (mmWave) radar, radar-based gait recognition has attracted increasing attention in recent years. In this research, we propose a Hierarchical Dynamic Network (HDNet) for gait recognition using mmWave radar. In order to explore more dynamic information, we propose point flow as a novel point clouds descriptor. We also devise a dynamic frame sampling module to promote the efficiency of computation without deteriorating performance noticeably. To prove the superiority of our methods, we perform extensive experiments on two public mmWave radar-based gait recognition datasets, and the results demonstrate that our model is superior to existing state-of-the-art methods.
\end{abstract}
\begin{keywords}
Millimeter-wave radar, gait recognition, point flow, hierarchical neural network, dynamic sampling
\end{keywords}
\section{Introduction}
\label{sec:intro}
Gait manifests a person's walking style, which is an important biometric \cite{sepas2022deep}. Recent efforts have been devoted to authenticating persons with their gaits~\cite{zhao2019mid,meng2020gait,cheng2021person,wang2022stpointgcn,ozturk2021gaitcube,marin2021ugaitnet,ni2020human,seifert2019toward,papanastasiou2021deep,cao2018radar}, i.e., gait recognition. Gain recognition has empowered a broad spectrum of applications, including personalized services, automatic surveillance, and human-computer interaction.
The rationale behind gait recognition is that individual gait consists of many different components (e.g., leg stride and arm swing) \cite{wan2018survey}, which can be used for individual identification. 


Recently, there is a trend of utilizing mmWave radar to implement gait recognition systems. Compared with wearable-based approaches, the mmWave-based approaches achieve non-contact gait recognition without discomforting users. Additionally, different from visual-based methods, the mmWave-based approach is more privacy-preserving and more robust to light and smoke conditions. Moreover, in contrast with other radio frequency-based approaches, e.g., Wi-Fi, the mmWave-based scheme is able to attain a higher spatial resolution due to its extremely high frequency (EHF) operating bands, and is well-suited for multi-target scenarios.


There have been some studies on mmWave-based gait recognition. 
As a pioneering work, mID~\cite{zhao2019mid} leverages a commodity mmWave radar to track and identify co-existing multi-people according to the point clouds captured by the radar. Later, a seminal dataset dedicated to mmWave-based gait recognition, along with a convolutional neural network (CNN)-based framework dubbed mmGaitNet~\cite{meng2020gait}. SPRNet \cite{cheng2021person} extends~\cite{meng2020gait} to the issue on person reidentification (ReID) by considering Doppler velocity and constructing a long-term sequence network. STPointGCN~\cite{wang2022stpointgcn} builds a spatial-temporal graph convolutional network (GCN) to achieve end-to-end gait recognition, as well as releases a new dataset. GaitCube~\cite{ozturk2021gaitcube} proposes an ingenious 3D aggregated feature representation called gait cube, which capacitates decent recognition accuracy even using minimal training data. 

However, all the above works only aggregate the spatiotemporal features in data frames but not investigating the information between frames. Intuitively, as walking is an active behavior, it is desirable to exploit the latent dynamic or periodic characteristics of human gaits.
To this end, we construct HDNet, a \underline{H}ierarchical \underline{D}ynamic \underline{Net}work, to efficiently aggregate the gait features and recognize gait. We propose a new radar point clouds feature descriptor, named \textit{point flow}, to extract the inter-frame information in point clouds. Moreover, we devise a Dynamic Frame Sampling (DFS) module to sample the informative radar point cloud frames to reduce the computation complexity.
We conduct extensive experiments to evaluate HDNet. Experimental results show that HDNet achieves 96.52\% and 91.4\% recognition accuracy on the mmGait dataset and STPointGCN dataset, which outperforms state-of-the-art works by 11.61\% and 15.89\%, respectively.
The contributions of our work can be summarized as:
\begin{itemize}[leftmargin=*,itemsep=0pt, topsep=0pt, partopsep=0pt, parsep=0.5pt]
    \item We propose a novel feature descriptor: point flow, which can dig more latent dynamic features of radar point clouds.
    \item We design a hierarchical network HDNet, which can aggregate spatial-temporal features of radar point clouds in a more efficient way.
    \item We devise a novel dynamic frame sampling module to sample point cloud data frames to improve the computation efficiency with marginal performance degradation.
\end{itemize}
The rest of the paper is organized as follows. \S\ref{sec:methodology} presents the design of HDNet. \S\ref{sec:experiment} evaluates the system performance and \S\ref{sec:conclusion} concludes the paper.

\section{Methodology}
\label{sec:methodology}
\subsection{Point Cloud and Point Flow}

\begin{figure}[t]
\centering
\includegraphics[width=\columnwidth]{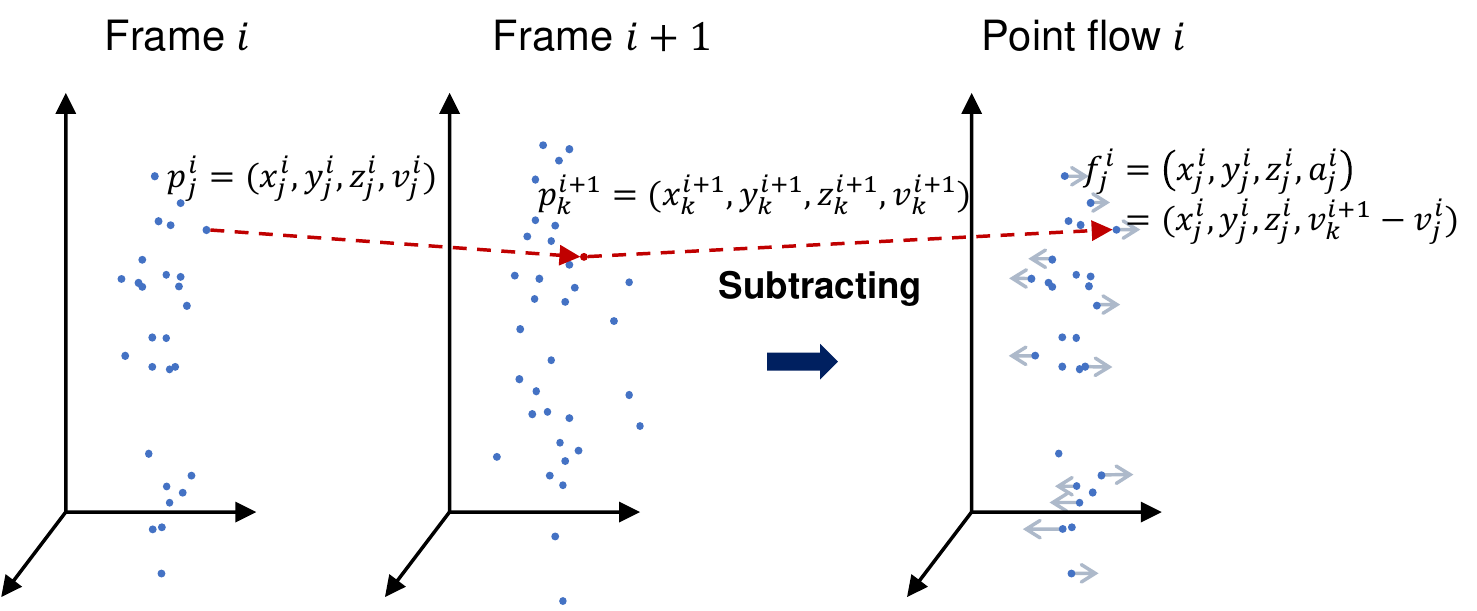} 
\caption{Point flow acquisition.}
\label{fig:pointflow}
\end{figure}

\begin{figure*}[t]
\centering
\includegraphics[width=\linewidth]{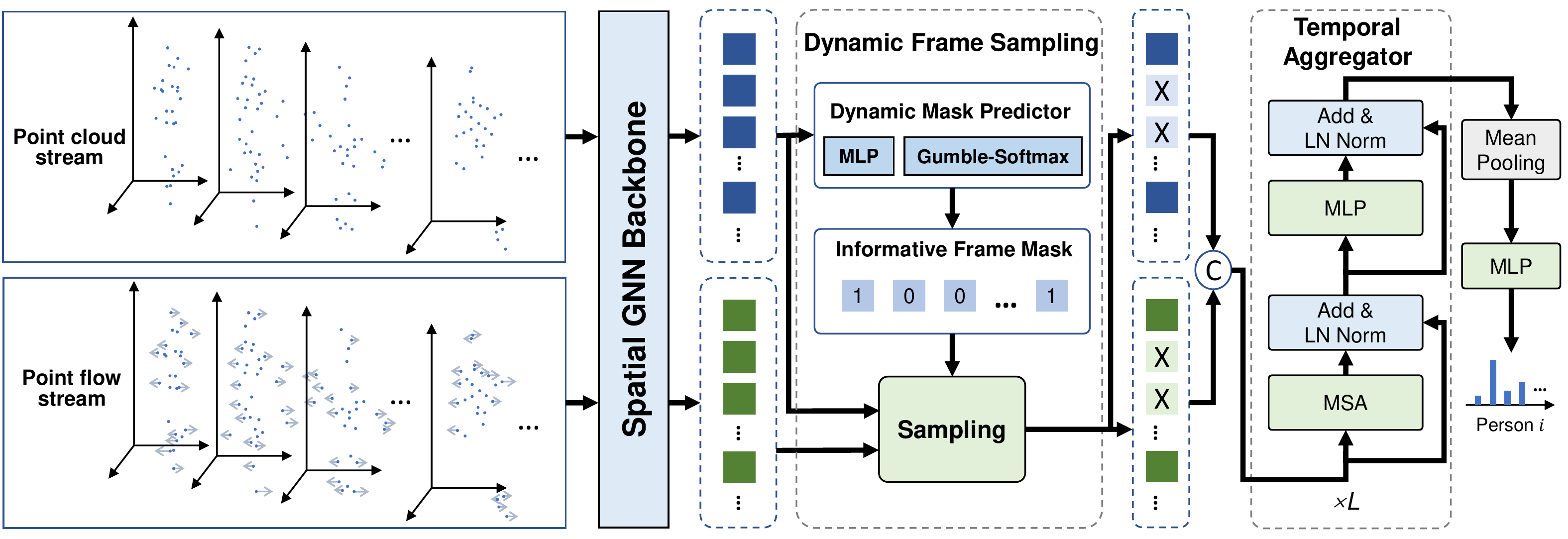} 
\caption{HDNet model structure.}
\label{model}
\vspace{-0.4cm}
\end{figure*}
The mmWave radar outputs the captured data in point clouds at a pre-defined sampling rate. We denote a series of radar point cloud frames with length $T$ as $S=\{P_0, P_1, ..., P_{T-1}\}$. Each frame contains $N$ points as $P_i=\{p^i_0, p^i_1, ..., p^i_{N-1}\}, i \in [0, T-1]$. For each point, it comprises four features $p^i_j=(x^i_j, y^i_j, z^i_j, v^i_j), j \in [0, N-1]$, where $x^i_j, y^i_j, z^i_j$ and $v^i_j$ represent 3D coordinates and Doppler velocity relative to radar, respectively. 

The mmWave radar returns the velocity information of each point according to the phase difference between chirps in a frame, which leverages the intra-frame information. Since the radar senses the target at a certain sampling rate, the information between two adjacent frames is missing by default. Thus, we propose \emph{point flow} to depict the dynamic changes of adjacent radar point cloud frames, which provides inter-frame information. As shown in Fig.~\ref{fig:pointflow}, for two adjacent frames $P_{i}$ and $P_{i+1}$, in order to get the relative changes between them, for point $p^{i}_j$ from frame $P_{i}$, we find the point $p^{i+1}_k$ closest to its spatial position in the next frame $P_{i+1}$, and by subtracting the Doppler velocity of these two points, we can get the corresponding point flow $f^i_j=(x^i_j, y^i_j, z^i_j, a^i_j)$, where $a^i_j = v^{i+1}_k - v^i_j$. For each radar point cloud frame $P_i$, we obtain the whole point flow as $F_i=\{f^i_0, f^i_1, ..., f^i_{N-1}\}$. Point flow provides more dynamic information from another dimension, which helps improve the recognition performance according to our experimental measurements (see \S\ref{subsec:ablation}).

\subsection{HDNet Structure}
Fig.~\ref{model} presents the structure of the proposed HDNet, which consists of three parts, i.e., Spatial Graph Neural Network (GNN) Backbone, Dynamic Frame Sampling (DFS), and Temporal Aggregator (TA).

\noindent \textbf{Spatial GNN Backbone.}
For succinctness, we refer to a sequence of radar point cloud frames $S$ as \emph{point cloud stream}, and the corresponding point flow frames as \emph{point flow stream}. First, for each frame in these two streams, we extract the frame-wise spatial features by using the Spatial GNN Backbone. Specifically, we leverage AdaptConv~\cite{zhou2021adaptive} as the backbone because it can capture the diverse relations between points from different semantic parts.

\noindent \textbf{Dynamic Frame Sampling.}
In practice, the mmWave radar may not cover the whole target. As a result, some point cloud frames only contain information about parts of the human body. To reduce computation workload and improve training efficiency, we design a DFS module to mask the uninformative frames and sample the informative frames dynamically. The informative point cloud and point flow are denoted by $\textbf{IP}^f$ and $\textbf{IP}^f$ respectively, which can be expressed by
\begin{equation}
    \begin{aligned}
        \textbf{IP}^f = {\rm Mask}({\rm AdaptConv}(P_i)),\ for \ i \in [0, T-1], \\
        \textbf{IF}^f = {\rm Mask}({\rm AdaptConv}(F_i)),\ for \ i \in [0, T-1].
    \end{aligned}
\end{equation}
To be specific, we define the informative frame mask $Mask\in \{0,1\}^N$  in which 1 means remaining and 0 refers to pruning. First, for frame $P_i$, we project the corresponding feature using an MLP network:
\begin{equation}
    \begin{aligned}
        \boldsymbol{z}_i &= {\rm MLP}({\rm AdaptConv}(P_i)) \\ 
        &= {\rm Linear}({\rm GeLU}({\rm LN}({\rm Linear}({\rm AdaptConv}(P_i))))) \in \mathbb{R}^{1 \times 2},
    \end{aligned}
\end{equation}
where ${\rm GeLU}$ is GeLU activation~\cite{hendrycks2016gaussian}, and ${\rm LN}$ represents layer normalization operation~\cite{ba2016layer}. Then, we get the probability of the frame remaining, denoted by $\pi$ using Softmax:
\begin{equation}
    \boldsymbol{\pi} = {\rm Softmax}(\boldsymbol{z}) \in \mathbb{R}^{M\times 2}.
\end{equation}
After that, we can generate the binary mask by sampling from $\boldsymbol{\pi}$.
The most common way to sample from $\boldsymbol{\pi}$ is simply applying the \emph{argmax} operation. However, \emph{argmax} is discrete and non-differentiable, which impedes end-to-end training in backpropagation.
To tackle this, we apply a reparameterization method~\cite{jang2017categorical}. In this way, the gradient of the non-differentiable sampling function is estimated from the Gumbel distribution during backpropagation. As such, we generate the discrete mask by using the Gumbel-Softmax technique:
\begin{equation}
    \begin{aligned}
        G_{i, j} &= {\rm Gumbel\text{-}Softmax}(\pi_{i, j})\\
        &= \frac{{\rm exp}(({\rm log}(\pi_{i, j}) + g_{i, j})/\tau)}{\sum^{1}_{k=0}{\rm exp}(({\rm log}(\pi_{i, j}) + g_{i, j})/\tau)}, {\rm for}\ j = 0, 1,
    \end{aligned}
\end{equation}
where $G_{i,j}$ are independent and identically sampled from $Gumbel(0,1)$ distribution, $\tau$ is the temperature hyperparameter. We obtain the binary mask by taking the first column of matrix $\boldsymbol{G}$, since the output of Gumbel-Softmax whose shape equals $\boldsymbol{\pi}$ is a one-hot tensor.
\begin{equation}
    Mask = G_{:, 0}\in \{0, 1\}^M.
\end{equation}

In order to facilitate training, we initialize all elements in the binary mask to 1 and update the mask progressively, since we find it enables the model to prune the uninformative frames gradually. In addition, to constrain the ratio of the remaining frames to a predefined value $t$, we add a mask loss to supervise the DFS module:
\begin{equation}
    Loss_{mask} = \left( t-\frac{{\rm sum}(Mask)}{{\rm len}(Mask)}\right)^2.
\end{equation}

\noindent \textbf{Temporal Aggregator.}
We concatenate the two informative features as $\textbf{I}={\rm Concat}(\textbf{IP}^f, \textbf{IF}^f)$, and feed them into the TA module to further aggregate the temporal features.
The TA module is a $L$-layer canonical Transformer~\cite{vaswani2017attention} block. Since gait features of human walking are periodic, which is suitable for Transformer networks, for can achieve great performance in modelling the interaction of features at long distances compared with Recurrent Neural Network (RNN) and Long Short-Term Memory (LSTM) network~\cite{hochreiter1997long}. The Transformer block consists of Multi-head Self-Attention (MSA), Multi-Layer Perceptron (MLP), and Layer Normalization (LN). The details are as follows:
\begin{equation}
    \begin{aligned}
        \textbf{I}^{l\prime\prime} &= \textbf{I}^{l-1} + {\rm MSA}({\rm LN} (\textbf{I}^{l-1})), \\
        \textbf{I}^{l\prime} &= \textbf{I}^{l\prime\prime} + {\rm MLP}({\rm LN},(\textbf{I}^{l\prime\prime}))
    \end{aligned}
\end{equation}
where $l=1,2,..,L$ is the index of the $l$-th TA block. Finally, we adopt mean-pooling for the output of the TA module, and get the final gait prediction by using MLP. 

The complete loss function can be denoted as:
\begin{equation}
    Loss = Loss_{NLLLoss} + \beta * Loss_{mask},
\end{equation}
where $Loss_{NLLLoss}$ is the Negative Log Likelihood Loss (NLLLoss) used to supervise the final prediction results, and $\beta$ is the weight of mask loss in the DFS module.

\section{Experiments}
\label{sec:experiment}
\subsection{Setup}
\noindent \textbf{Dataset.}
We conduct extensive experiments on two public datasets for human gait recognition.

\begin{itemize}[leftmargin=*,itemsep=0pt, topsep=0pt, partopsep=0pt, parsep=0.5pt]
\item \textbf{mmGait} 
\cite{meng2020gait} collects a total of 30 hours of 3D point cloud data from 95 volunteers using TI IWR1443
and TI IWR6843
as the mmWave sensing devices. It contains two types of walking trajectories, i.e., fixed route and free route, with up to 5 volunteers walking simultaneously.

\item \textbf{STPointGCN}
\cite{wang2022stpointgcn} contains a total of 75 minutes of 3D point cloud data from 6 participants in three different environments using TI IWR6843. 
\end{itemize}

\noindent \textbf{Data Preprocessing.}
For the mmGait dataset, we first utilize the DBSCAN clustering algorithm to divide the points in a frame into different groups/persons. Then, we track the clustered point clouds using the Hungarian algorithm to get the continuous gait data of persons. Note that this process for separating co-existing people introduces recognition error. To fairly compare the performance of gait recognition models,
we select the radar data that only has one person walking. For the STPointGCN dataset, since it already separates gait data by using the built-in clustering and tracking algorithm, there is no need for any clustering and tracking procedure. 

Given the number of radar points differs among frames, and to exclude the noisy frames, we discard the frames whose number of points is less than $N$=16. Additionally, a non-overlapped sliding window with window length $T$=20 as the moving step, is used to generate data that can adapt the input of the model. Besides, in order to form the data as a batch of samples, we sample $N$ points from the original point cloud for each frame by adopting the furthest point sampling method as PointNet++~\cite{qi2017pointnet++}.

\noindent \textbf{Metrics.} The metrics we used in mmGait dataset are accuracy, precision, and F1-score, and we also report the experiment results after 5-fold cross-validation. For STPointGCN dataset, we report the precision, recall, and F1-score for ease of comparison with the original results reported in~\cite{wang2022stpointgcn}. 

\subsection{Experiment Results}
\begin{table}[t]
\centering
\caption{Results on mmGait dataset.}
\setlength{\tabcolsep}{3.8mm}
\label{mmgait_result}
\begin{threeparttable}
\resizebox{\linewidth}{!}{
\begin{tabular}{lllll}
\toprule
Method      & Accuracy & Precision & F1-score \\  \midrule
mmGaitNet   & 39.10±1.15 & 38.74±1.99 & 37.34±1.76 \\
PL          & 80.77±3.15 & 81.52±4.81 & 79.91±3.28 \\
P+L         & 81.23±2.98 & 80.84±3.25 & 80.84±2.84 \\
SRPNet      & 84.91±2.86 & 85.26±2.47 & 84.67±2.18 \\
\midrule
\textbf{HDNet (ours)} & \textbf{96.52±1.26} & \textbf{96.48±1.32} & \textbf{96.36±1.37} \\
\bottomrule
\end{tabular}
}
\end{threeparttable}
\vspace{-0.4cm}
\end{table}

\begin{table}[t]
\centering
\caption{Results on STPointGCN dataset.}
\setlength{\tabcolsep}{3.8mm}
\label{stpointgcn_result}
\begin{threeparttable}
\begin{tabular}{llll}
\toprule
Method      & Precision & Recall & F1-score \\  \midrule
PL          & 67.71    & 80.71 & 73.09   \\
P+L         & 65.23    & 77.88 & 70.38   \\
PGL         & 65.99    & 65.07 & 65.31   \\
mmGaitNet   & 70.48    & 72.48 & 70.84   \\
STPointGCN  & 75.51    & 79.22 & 76.50   \\
\midrule
\textbf{HDNet (Ours)} & \textbf{91.40} & \textbf{91.42} & \textbf{91.29} \\
\bottomrule
\end{tabular}
\end{threeparttable}
\vspace{-0.4cm}
\end{table}
We first compare our proposed HDNet with other models without any frame sampling.
For mmGait dataset, we evaluate different benchmark algorithms including PointNet~\cite{qi2017pointnet} combined with LSTM (PL), PointNet++~\cite{qi2017pointnet++} combined with LSTM (P+L), mmGaitNet~\cite{meng2020gait}, and SRPNet~\cite{cheng2021person}. As shown in Table~\ref{mmgait_result}, mmGaitNet performs significantly worse than PL and P+L in that we only sample $N$=16 points for each frame, which restricts the CNN to aggregate the spatiotemporal features. Notably, our proposed HDNet achieves 13.7\% improvement over SOTA model SRPNet in terms of accuracy.
As for STPointGCN dataset, we keep to the same experiment setup as~\cite{wang2022stpointgcn}. The experimental results are provided in Table~\ref{stpointgcn_result}, which indicates our proposed HDNet outperforms other models. In particular, the proposed model capacitates 21\% improvement over SOTA model STPointGCN in accuracy.

\begin{figure}[t]
\centering
\includegraphics[width=0.9\columnwidth]{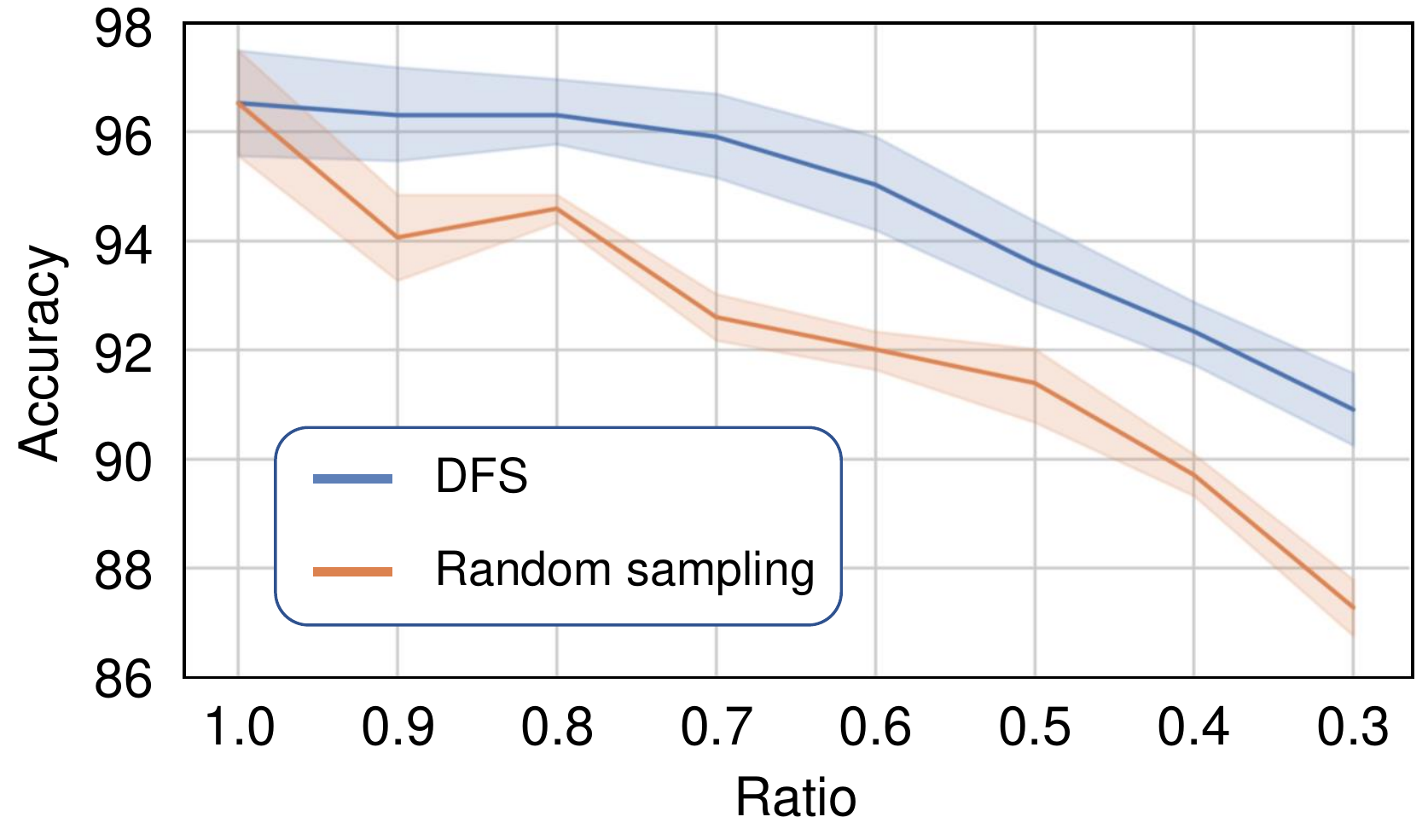} 
\vspace{-0.4cm}
\caption{Model performance with different sampling methods across different sampling ratios.}
\label{ratio}
\vspace{-0.2cm}
\end{figure}

Next, we conduct experiments to explore the effectiveness of our HDNet in mmGait dataset across different sampling ratio, as well as demonstrate the superiority of our DFS module compared to random sampling, which is just sampling the frames randomly under the given ratio of sampling. The overall results are presented in Fig.~\ref{ratio}. It can be noticed that the accuracy of both methods declines as the ratio of the remaining frames (sampling ratio) decreases. On the flip side, compared with using random sampling, our HDNet is less vulnerable to the sampling ratio by using DFS module. Even when the sampling ratio drops to 0.3, the accuracy obtained by using DFS module still remains above 90\%, which is remarkably higher than using random sampling.

\subsection{Ablation Study}
\label{subsec:ablation}
Finally, we proceed to perform experiments in light of analyzing the effect of each module in our HDNet model, and the results are summarized in Table~\ref{ablation}. We first conduct experiments on HDNet without spatial GNN backbone, i.e., replace GNN with PointNet, and the performance degrades dramatically, which pinpoints that GNN is suitable for feature extraction of radar point clouds. We then substitute LSTM for the Transformer-based TA module, and the performance also declines, which reveals Transformer block can aggregate radar temporal gait features more efficiently than LSTM. Furthermore, we investigate the performance of HDNet without exploiting point flow features. It is evident that all the metrics also decay, which demonstrates that our proposed point flow is also a paramount feature in mining the dynamic characteristic of temporal radar point cloud frames.

\begin{table}[]
\footnotesize
\centering
\caption{Ablation study.}
\label{ablation}
\begin{threeparttable}
\begin{tabular}{lllll}
\toprule
Method      & Accuracy & Precision & F1-score \\  \midrule
HDNet w/o GNN           & 81.50±0.92 & 81.72±0.87 & 81.02±0.95    \\
HDNet w/o TA module   & 88.56±0.83 & 88.27±0.94 & 88.16±0.94 \\
HDNet w/o point flow    & 92.18±0.95 & 91.89±0.93 & 91.88±0.87   \\
\midrule
\textbf{HDNet} & \textbf{96.52±1.26} & \textbf{96.48±1.32} & \textbf{96.36±1.37} \\
\bottomrule
\end{tabular}
\end{threeparttable}
\vspace{-0.4cm}
\end{table}

\section{Conclusion}
\label{sec:conclusion}
In this paper, we proposed a novel radar point feature descriptor dubbed point flow, which can explore more dynamic characteristics of radar point clouds. Besides, we constructed a novel hierarchical dynamic network for gait recognition named HDNet, which can aggregate the spatial and temporal features of human gait data more efficiently. Moreover, in order to promote the efficiency of our model, we devised an ingenious dynamic frame sampling module to sample the informative frames adaptively. 
\balance
\bibliographystyle{IEEEbib}
\bibliography{refs}

\end{document}